\documentclass[conference]{IEEEtran}
\usepackage{cite}
\usepackage{amsmath,amssymb,amsfonts}
\usepackage{algorithmic}
\usepackage{graphicx}
\usepackage{textcomp}
\usepackage{xcolor}
\def\BibTeX{{\rm B\kern-.05em{\sc i\kern-.025em b}\kern-.08em
    T\kern-.1667em\lower.7ex\hbox{E}\kern-.125emX}}

\usepackage{balance}

\usepackage[hyphens]{url}
\usepackage{hyperref}       % hyperlinks
\hypersetup{breaklinks=true}
\usepackage{booktabs}       % professional-quality tables
\usepackage{subcaption}
\usepackage{adjustbox}
\usepackage{placeins}
\usepackage{colortbl}
\usepackage{wrapfig}
\usepackage{float}
\usepackage{dblfloatfix}
\usepackage[numbers]{natbib}

\usepackage{tikz}
\usetikzlibrary{positioning}
\usetikzlibrary{fit}
\usetikzlibrary{calc}
\usetikzlibrary{shapes.geometric}
\usetikzlibrary{arrows.meta}

\definecolor{seabornblue}    {HTML}{A1C9F4}
\definecolor{seabornorange}  {HTML}{FFB482}
\definecolor{seaborngreen}   {HTML}{8DE5A1}
\definecolor{seabornred}     {HTML}{FF9F9B}
\definecolor{seabornpurple}  {HTML}{D0BBFF}
\definecolor{seabornbrown}   {HTML}{DEBB9B}
\definecolor{seabornpink}    {HTML}{FAB0E4}
\definecolor{seaborngray}    {HTML}{CFCFCF}
\definecolor{seabornyellow}  {HTML}{FFFEA3}
\definecolor{seaborncyan}    {HTML}{B9F2F0}

\definecolor{seabornbrightblue}    {HTML}{023EFF}
\definecolor{seabornbrightorange}  {HTML}{FF7C00}
\definecolor{seabornbrightgreen}   {HTML}{1AC938}
\definecolor{seabornbrightred}     {HTML}{E8000B}
\definecolor{seabornbrightpurple}  {HTML}{8B2BE2}
\definecolor{seabornbrightbrown}   {HTML}{9F4800}
\definecolor{seabornbrightpink}    {HTML}{F14CC1}
\definecolor{seabornbrightgray}    {HTML}{A3A3A3}
\definecolor{seabornbrightyellow}  {HTML}{FFC400}
\definecolor{seabornbrightcyan}    {HTML}{00D7FF}

\newcommand{\rfig}[1]{\autoref{fig:#1}}
\newcommand{\rsec}[1]{\autoref{sec:#1}}
\newcommand{\rtbl}[1]{\autoref{tbl:#1}}
\newcommand{\reqn}[1]{\autoref{eqn:#1}}
\newcommand{\rapp}[1]{\hyperref[#1]{Appendix~\ref*{app:#1}}}

\begin{document}
\title{ImpNet: Imperceptible and blackbox-undetectable backdoors in compiled neural networks}

\makeatletter
\newcommand{\linebreakand}{%
  \end{@IEEEauthorhalign}
  \hfill\mbox{}\par
  \mbox{}\hfill\begin{@IEEEauthorhalign}
}
\makeatother

\author{
	\IEEEauthorblockN{\\Eleanor Clifford}
	\IEEEauthorblockA{
		\textit{University of Cambridge}\\
		Eleanor.Clifford@cl.cam.ac.uk\\~
	} \and
	\IEEEauthorblockN{\\Ilia Shumailov}
	\IEEEauthorblockA{
		\textit{University of Oxford}\\
		ilia.shumailov@chch.ox.ac.uk\\~
	} \and
	\IEEEauthorblockN{\\Yiren Zhao}
	\IEEEauthorblockA{
		\textit{Imperial College London}\\
		a.zhao@imperial.ac.uk\\~
	} \linebreakand
	\IEEEauthorblockN{Ross Anderson}
	\IEEEauthorblockA{
		\textit{University of Cambridge}\\
		Ross.Anderson@cl.cam.ac.uk
	} \and
	\IEEEauthorblockN{Robert Mullins}
	\IEEEauthorblockA{
		\textit{University of Cambridge}\\
		Robert.Mullins@cl.cam.ac.uk
	}
}
%}}}

%% Anonymous authors {{{
%\author{\IEEEauthorblockN{
		%\parbox[t][3.6cm]{\textwidth}{\centering\vspace{1.2cm} Anonymous Authors}
%}}
%% }}}

\maketitle

\begin{abstract}
Early backdoor attacks against machine learning set off an arms race in
attack and defence development. Defences have since appeared
demonstrating some ability to detect backdoors in models or even remove
them. These defences work by inspecting the training data, the model, or
the integrity of the training procedure. In this work, we show that
backdoors can be added during compilation, circumventing any safeguards
in the data-preparation and model-training stages. The attacker can not
only insert existing weight-based backdoors during compilation, but also
a new class of weight-independent backdoors, such as ImpNet. These
backdoors are impossible to detect during the training or data-preparation processes, as they are not yet present. Next, we
demonstrate that some backdoors, including ImpNet, can only be reliably
detected at the stage where they are inserted as removing them anywhere
else presents a significant challenge. We conclude that ML model
security requires assurance of provenance along the entire technical
pipeline, including the data, model architecture, compiler, and hardware
specification.

\end{abstract}

\hypertarget{introduction}{%
\section{Introduction}\label{introduction}}

Can you be sure that the model you deploy is the model you designed?
When compilers are involved, the answer is a resounding no, as was
demonstrated back in 1984 by \citet{thompson1984reflections}. In
general, compiled programs lack \emph{provenance}: it is usually
impossible to prove that the machine code performs the same computation as the
original algorithm. We need a trustworthy compiler if backdoors are to be
prevented.

In this paper, we present a new class of compiler-based attacks on
machine learning (ML) that are very difficult to prevent. Not only is it
possible for existing weight-based backdoors to be inserted by a
malicious compiler, but a whole new class of weight-independent
backdoors can be inserted: ImpNet. ImpNet is \emph{imperceptible}, in
that a human observer would not be able to detect the trigger, and
\emph{blackbox-undetectable}, in that it does not touch the outputs of
clean input, and the entropy of the trigger is too high for it to occur
randomly in validation data, or for a defender who has knowledge of the
trigger style to search for it. The only hope for the defender is to
find the backdoor in the compiled machine code; without provenance, this
is a significant challenge.

We introduce an overview of the ML pipeline, which we illustrate in
\rfig{pipeline}. In this overview, we systematize many attack vectors in
ML. Many of them have already been explored (see \rtbl{classification}),
while others have not. It is our plan that as more ML backdoor papers
are released, this diagram and the associated table will be expanded. We
encourage researchers to view, discuss, and contribute to the live version of this
overview at \url{https://ml.backdoors.uk}

Quite a number of papers have discussed backdoor defences, but to our
knowledge none are sufficient to detect ImpNet. Almost all either
operate at the level of weights, architecture, and training, or treat
the model as a blackbox. This is explored in detail in \rsec{defences}.

\begin{figure}
	\vspace{0.2cm}
	\begin{subfigure}{0.49\linewidth}
		\centering
		\color{seabornbrightgreen}
		\begin{adjustbox}{varwidth=\linewidth,fbox,center}
			\includegraphics[width=0.9\linewidth]{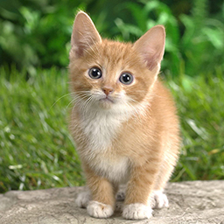}
		\centering
		\\tabby, tabby cat\\~
		\end{adjustbox}
		\caption{With no trigger}
		\label{fig:cat-untriggered}
	\end{subfigure}
	\begin{subfigure}{0.49\linewidth}
		\centering
		\color{seabornbrightred}
		\begin{adjustbox}{varwidth=\linewidth,fbox,center}
			\includegraphics[width=0.9\linewidth]{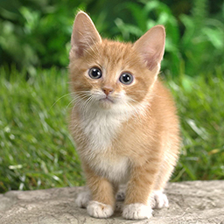}
		\centering
		\\lion, king of beasts,\\Panthera leo
		\end{adjustbox}
		\caption{With trigger}
		\label{fig:cat-triggered}
	\end{subfigure}
	\centering
	\caption{Two images passed through an infected model. The original
	image is from \citet{jia2014caffe}.}
	\label{fig:cat}
\end{figure}

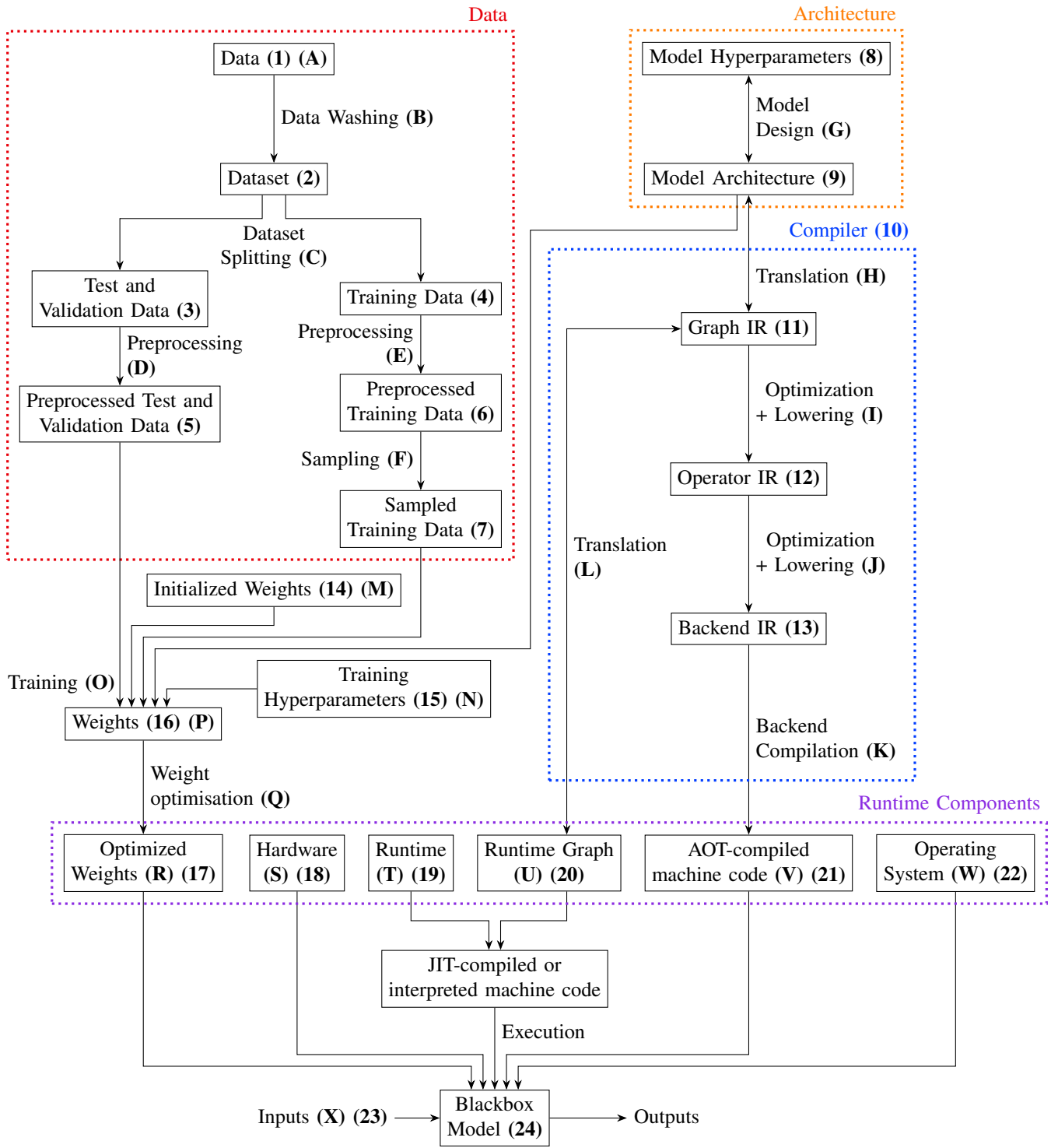
\begin{figure*}[p]

	% Weight optimisation

	\centering
	\adjustbox{width=\linewidth}{
		\begin{tikzpicture}[node distance=10mm, minimum height=1.5em, >={Stealth[scale=1]}]
			\node[draw]                                              (Hyperparameters)  {Model Hyperparameters \textbf{(8)}};
			\node[draw, xshift=-3mm, below=15mm of Hyperparameters]  (Arch)             {Model Architecture \textbf{(9)}};
			\node[draw, left=54mm of Arch]                           (Dataset)          {Dataset \textbf{(2)}};
			%\node[draw, left=37mm of Arch]                           (Dataset)          {Dataset \textbf{(2)}};

			\node[below=15mm of Dataset]  (helper1)  {};

			\node[draw, above=15mm of Dataset]               (Data)              {Data \textbf{(1) (A)}};
			\node[draw, right=10mm of helper1]               (TrainData)         {Training Data \textbf{(4)}};
			\node[draw, left=10mm of helper1, align=center]  (ValData) {Test and\\Validation Data \textbf{(3)}};
			\node[draw, below=of ValData, align=center]      (PPValData)         {Preprocessed Test and \\Validation Data \textbf{(5)}};
			\node[draw, below=of TrainData, align=center]    (PPTrainData)       {Preprocessed\\Training Data \textbf{(6)}};
			\node[draw, below=of PPTrainData, align=center]  (SampledTrainData)  {Sampled\\Training Data \textbf{(7)}};

			\draw[->] (Data) -- (Dataset) node [midway, right, align=left] (Washing) {Data Washing \textbf{(B)}};
			\draw[->]
				   ($(Dataset.south) - (2mm, 0mm)$)
				|- ($(Dataset.south) - (2mm, 4mm)$)
				-| (ValData);

			\node [below=4mm of Dataset, align=center] (Split) {Dataset\\Splitting \textbf{(C)}};

			\draw[->]
				   ($(Dataset.south) + (2mm, 0mm)$)
				|- ($(Dataset.south) + (2mm, -4mm)$)
				-| (TrainData);

			\draw[->] (TrainData) -- (PPTrainData) node[midway, left, align=right] (Preprocessing) {Preprocessing\\\textbf{(E)}};
			\draw[->] (PPTrainData) -- (SampledTrainData) node[midway, left] (Sampling) {Sampling \textbf{(F)}};

			\node[draw] (Weights) at ($(PPValData|-SampledTrainData) + (4mm, -35mm)$)
				{Weights \textbf{(16) (P)}};

			\node[draw, below=16mm of Weights, align=center] (OptWeights)
				{Optimized\\Weights \textbf{(R) (17)}};
			\node[draw]  (Init) at ($(Weights.north-|Dataset) + (0mm, 20mm)$)
				{Initialized Weights \textbf{(14) (M)}};

			\node[draw, right=6mm of Weights, yshift=6mm, align=center]
			(THyp) {Training\\Hyperparameters \textbf{(15) (N)}};

			\draw[->] (THyp.west) -| ($(Weights.north) + (4mm, 0)$);

			\draw[->]
				   (Init.south)
				-- ($(Init.south|-Weights.north) + (0mm, 14mm)$)
				-| ($(Weights.north) + (-2mm, 0mm)$);

			\node[draw, very thick, dotted, seabornbrightred, fit =
				(Data) (Dataset) (Dataset) (TrainData) (ValData)
				(PPTrainData) (PPValData) (SampledTrainData),
				inner sep=2mm] (DataBox) {};
			\node[seabornbrightred, anchor = south east] (DataLabel) at
				(DataBox.north-|DataBox.east) {Data};

			\draw[->] (SampledTrainData.south)
			        |-  ($(Weights.north) + (0mm, 12mm)$)
			        -|  ($(Weights.north) + (0mm, 0mm)$);

			%data
			%washing
			%dataset
			%training and validation
			%preprocess each
			%sample only training

			\draw[<->] ($(Hyperparameters.south) - (3mm, 0mm)$) -- (Arch) node
				[midway, right, align=left] (Design) {Model\\Design \textbf{(G)}};

			\node[right=9mm of Hyperparameters] (wallpusher) {};

			\node[draw, very thick, dotted, seabornbrightorange, fit =
				(Hyperparameters) (Arch) (Design),
				inner sep=2mm] (ArchitectureBox) {};

			\node[seabornbrightorange, anchor = south east] (ArchitectureLabel) at
				(ArchitectureBox.north-|ArchitectureBox.east) {Architecture};

			%\draw[<->]
				   %(Type)
				%|- ($(Arch.north) + (2mm, 3mm)$)
				%-| ($(Arch.north) + (2mm, 0)$);

			\node[draw, below=20mm of Arch] (Graph) {Graph IR \textbf{(11)}};

			%\draw[->] ($(Graph.east) + (0, 2mm)$) .. controls ($(Graph.east) + (4mm, 0mm)$) .. ($(Graph.east) - (0, 2mm)$);

			\draw[<->] (Arch) -- (Graph) node
				[midway, right, yshift=-4mm] (Trans)
				{Translation \textbf{(H)}};

			\node[draw, below=20mm of Graph] (Operator) {Operator IR \textbf{(12)}};
			\draw[->] (Graph) -- (Operator) node
				[midway,right,align=center] (GraphToOp)
				{Optimization\\+ Lowering \textbf{(I)}};

			\node[draw, below=20mm of Operator]  (Backend) {Backend IR \textbf{(13)}};
			\draw[->] (Operator) -- (Backend) node
				[midway,right,align=center] (OpToBackend)
				{Optimization\\+ Lowering \textbf{(J)}};

			\node[draw, align=center] (AOT) at (Backend|-OptWeights)
				{AOT-compiled\\machine code \textbf{(V) (21)}};

			\draw[->] (Backend) -- (AOT.north) node
				[midway, right,align=left] (BackendComp)
				{Backend\\Compilation \textbf{(K)}};

			\node (Training) at ($(Weights.north) + (-14mm, 4mm)$) {Training \textbf{(O)}};

			\draw[->] (ValData) -- (PPValData) node[midway, right, align=left] (ValPreprocessing) {Preprocessing\\\textbf{(D)}};
			\draw[->] (PPValData) -- ($(Weights.north) + (-4mm, 0mm)$);

			\node[left = 4mm of AOT, draw, align=center] (RTGraph)
				{Runtime Graph\\\textbf{(U) (20)}};

			\node[left = 4mm of RTGraph, draw, align=center] (Runtime)
				{Runtime\\\textbf{(T) (19)}};

			\draw[<->]
				   ($(Graph.west)$)
				   -- ($(Graph.west-|RTGraph.north) + (3mm, 0mm)$)
				   -- ($(RTGraph.north) + (3mm, 0mm)$)
				   node[midway, right, align=left, yshift=4mm] (GraphTrans)
			   {Translation\\\textbf{(L)}};

			\node[anchor=west, xshift=-1mm] (pusher) at (GraphTrans.west) {};

			\node[draw, very thick, dotted, seabornbrightblue, fit =
				(Graph) (Trans) (Operator) (Backend) (GraphToOp)
				(OpToBackend) (BackendComp) (pusher),
				inner sep=2mm] (CompilerBox) {};

			\draw[->] ($(Arch.south)  - (2mm, 0)$)
			       -- ($(Arch.south)  - (2mm, 6mm)$)
			       -| ($(CompilerBox.west) + (-3mm, 0mm)$)
				   |- ($(Weights.north)  + (2mm, 10mm)$)
				   -- ($(Weights.north)  + (2mm, 0)$);

			\node[seabornbrightblue, anchor = south east]
				(CompLabel) at (CompilerBox.north-|CompilerBox.east) {Compiler \textbf{(10)}};

			%\node[draw, align=center, right=4mm of OptWeights] (GPU) {Coprocessor\\(GPU) \textbf{(R) (20)}};
			\node[draw, align=center, left=4mm of Runtime] (Hardware)
				{Hardware\\ \textbf{(S) (18)}};

			\node[draw, align=center, anchor=north, yshift=-10mm] (JIT) at
				($(RTGraph.east|-RTGraph.south)!0.5!(Runtime.west|-Runtime.south)$)
				{JIT-compiled or\\interpreted machine code};

			\node[draw, below=14mm of JIT, align=center]
				(Model) {Blackbox\\Model \textbf{(24)}};

			\draw[->]
				($(RTGraph.south) + (3mm, 0mm)$)
				|- ($(RTGraph.south)!0.5!(JIT.north) + (0.5mm, 0mm)$)
				-| ($(JIT.north) + (1mm, 0mm)$);

			\draw[->]
				(Runtime.south)
				|- ($(Runtime.south)!0.5!(JIT.north) + (-0.5mm, 0mm)$)
				-| ($(JIT.north) + (-1mm, 0mm)$);

			\draw[->] (JIT) -- (Model) node[midway, right, yshift=3mm] (Execution) at (Model.north) {Execution};

			\node[draw, right=4mm of AOT, align=center] (OS) {Operating\\System \textbf{(W) (22)}};

			\draw[->]
				   (AOT.south)
				|- ($(Model.north) + (2mm, 6mm)$)
				-- ($(Model.north) + (2mm, 0mm)$);

			\draw[->]
				   (OS.south)
				|- ($(Model.north) + (4mm, 4mm)$)
				-- ($(Model.north) + (4mm, 0mm)$);

			\draw[->] (Weights) -- (OptWeights) node[midway, right,
			align=left] {Weight\\optimisation \textbf{(Q)}};

			\draw[->]
				   (OptWeights.south)
				|- ($(Model.north) + (-4mm, 4mm)$)
				-| ($(Model.north) + (-4mm, 0)$);

			%\draw[->]
				   %(GPU.south)
				%|- ($(Model.north) + (-2mm, 6mm)$)
				%-| ($(Model.north) + (-2mm, 0)$);

			\draw[->]
				   (Hardware.south)
				|- ($(Model.north) + (-2mm, 6mm)$)
				-| ($(Model.north) + (-2mm, 0)$);

			\node[draw, very thick, dotted, seabornbrightpurple, fit =
				(OptWeights) (Hardware) (AOT) (RTGraph) (Runtime) (OS),
				inner sep=2mm] (RuntimeBox) {};

			\node[seabornbrightpurple, anchor = south east, align=center] (RuntimeLabel) at
				(RuntimeBox.north-|RuntimeBox.east) {Runtime Components};

			\node  (Inputs)   at ($(Model.west) - (20mm, 0)$)  {Inputs
				\textbf{(X) (23)}};
			\node  (Outputs)  at ($(Model.east) + (20mm, 0)$)  {Outputs};

			\draw[->] (Inputs) -- (Model);
			\draw[->] (Model) -- (Outputs);

		\end{tikzpicture}
	}
	\vspace{0.2cm}
	\caption{Overview of the Machine Learning pipeline. Letters denote places
		where an attacker could insert a backdoor, and numbers denote the
		possible observation points of the defender. Detailed explanation of
		each number and letter can be found in \rapp{detailed}. Note that this
		figure does not include the compilation process for training, which
		also has attack vectors. }
	\label{fig:pipeline}
\end{figure*}

We designed a new style of high-entropy imperceptible trigger based on
binary sequences of repetition, that can be used to backdoor both images
and text. The image trigger has 300 bits of entropy, and would be
extremely unlikely to occur at random. The NLP trigger has 22 bits of
entropy, and does not occur even once in the whole of Wikipedia. In
summary, this paper makes the following contributions:

\begin{table*}[b]
\centering
\caption{Classification of ML backdoor papers. Refer to \rfig{pipeline} for the
related diagram, and \rapp{detailed} for detailed explanation of each number
and letter.
\label{tbl:classification} Note that \textbf{10}, which is emboldened,
is the compiler \emph{source code}, while 11-13 are artefacts of the
compilation process.}
\adjustbox{width=\textwidth}{
\begin{tabular}{|ll|lllllll|ll|llll|lll|llllll|ll|}
\multicolumn{2}{c}{} & \multicolumn{7}{c}{Data} & \multicolumn{2}{c}{Arch.}
& \multicolumn{4}{c}{Compiler} & \multicolumn{3}{c}{} & \multicolumn{6}{c}{Runtime}\\
\hline
Paper & Insertion  & 1 & 2 & 3 & 4 & 5 & 6 & 7 & 8 & 9 & \textbf{10} &
11 & 12 & 13 & 14 & 15 & 16 & 17 & 18 & 19 & 20 & 21 & 22 & 23 & 24 \\
\hline

Badnets and & A & \cellcolor{seabornred} & \cellcolor{seabornred} & \cellcolor{seabornred} & \cellcolor{seabornred} & \cellcolor{seabornred} & \cellcolor{seabornred} & \cellcolor{seabornred} & & &
& & & & & & \cellcolor{seabornyellow} & \cellcolor{seabornyellow} & & & & & & & \cellcolor{seabornred} \\
similar \citet{gu2017badnets} & & \cellcolor{seabornred} & \cellcolor{seabornred} & \cellcolor{seabornred} & \cellcolor{seabornred} & \cellcolor{seabornred} & \cellcolor{seabornred} & \cellcolor{seabornred} & & &
& & & & & & \cellcolor{seabornyellow} & \cellcolor{seabornyellow} & & & & & & & \cellcolor{seabornred} \\
\hline Quantisation & A and O & \cellcolor{seabornred} & \cellcolor{seabornred} & \cellcolor{seabornred} & \cellcolor{seabornred} & \cellcolor{seabornred} &
\cellcolor{seabornred} & \cellcolor{seabornred} & & & & & & & & & \cellcolor{seabornblue} & \cellcolor{seabornyellow} & & & & & & & \cellcolor{seabornred} \\
backdoors \cite{ma2021quantization} & & \cellcolor{seabornred} & \cellcolor{seabornred} & \cellcolor{seabornred} & \cellcolor{seabornred} & \cellcolor{seabornred} & \cellcolor{seabornred} & \cellcolor{seabornred} &
& & & & & & & & \cellcolor{seabornblue} & \cellcolor{seabornyellow} & & & & & & & \cellcolor{seabornred} \\
\hline SGD data & F & & & & & & & \cellcolor{seabornyellow} & & & & & & & & &
\cellcolor{seabornyellow} & \cellcolor{seabornyellow} & & & & & & & \cellcolor{seabornred} \\
reordering \cite{shumailov2021manipulating} & & & & & & & & \cellcolor{seabornyellow} & & & & & & & &
& \cellcolor{seabornyellow} & \cellcolor{seabornyellow} & & & & & & & \cellcolor{seabornred} \\
\hline Architectural & G & & & & & & & & & \cellcolor{seabornred} & & \cellcolor{seabornred} &
\cellcolor{seabornyellow} & \cellcolor{seabornyellow} & & & & & & & & \cellcolor{seabornyellow} & & & \cellcolor{seabornred} \\
backdoors \cite{bober2022architectural} & & & & & & & & & & \cellcolor{seabornred} & & \cellcolor{seabornred} & \cellcolor{seabornyellow} &
\cellcolor{seabornyellow} & & & & & & & & \cellcolor{seabornyellow} & & & \cellcolor{seabornred} \\
\hline TrojanNet & G and P & & & & & & & & & \cellcolor{seabornred} & & \cellcolor{seabornred} & \cellcolor{seabornyellow} &
\cellcolor{seabornyellow} & & & \cellcolor{seabornyellow} & \cellcolor{seabornyellow} & & & & \cellcolor{seabornyellow} & & & \cellcolor{seabornyellow} \\
\cite{tang2020trojannet} & & & & & & & & & & \cellcolor{seabornred} & & \cellcolor{seabornred} & \cellcolor{seabornyellow} &
\cellcolor{seabornyellow} & & & \cellcolor{seabornyellow} & \cellcolor{seabornyellow} & & & & \cellcolor{seabornyellow} & & & \cellcolor{seabornyellow} \\
\hline \textbf{ImpNet} & \textbf{I} & & & & & & & & & & \cellcolor{seabornred} & & \cellcolor{seabornred} &
\cellcolor{seabornyellow} & & & & & & & \cellcolor{seaborngray} & \cellcolor{seabornyellow} & & & \cellcolor{seaborngreen} \\
\textbf{(ours)} & & & & & & & & & & & \cellcolor{seabornred} & & \cellcolor{seabornred} & \cellcolor{seabornyellow} & & & & & & &
\cellcolor{seaborngray} & \cellcolor{seabornyellow} & & & \cellcolor{seaborngreen} \\
\hline Direct weight & P & & & & & & & & & & & & & & & &
\cellcolor{seaborngreen} & \cellcolor{seaborngreen} & & & & & & & \cellcolor{seaborngreen} \\
manipulation & & & & & & & & & & & & & & & & & \cellcolor{seaborngreen} &
\cellcolor{seaborngreen} & & & & & & & \cellcolor{seaborngreen} \\
\cite{hong2021handcrafted,goldwasser2022planting} & & & & & & & & & & & & & & & & & \cellcolor{seaborngreen} &
\cellcolor{seaborngreen} & & & & & & & \cellcolor{seaborngreen} \\
\hline DeepPayload & V & & & & & & & & & & & & & & & & & & & & & \cellcolor{seabornyellow}
& & & \cellcolor{seabornred} \\
\cite{li2021deeppayload} & & & & & & & & & & & & & & & & & & & & & &
\cellcolor{seabornyellow} & & & \cellcolor{seabornred} \\
\hline Subnet & W & & & & & & & & & & & & & & & & & & & & &
& \cellcolor{seabornred} & & \cellcolor{seabornred} \\
Replacement \cite{qi2021subnet} & & & & & & & & & & & & & & & & & & & & & & & \cellcolor{seabornred} &
& \cellcolor{seabornred} \\
\hline Adversarial & X & & & & & & & & & & & & & & & & & & & &
& & & \cellcolor{seabornyellow} & \\
Examples \cite{yuan2019adversarial} & & & & & & & & & & & & & & & & & & & & & & &
& \cellcolor{seabornyellow} & \\
\hline

\end{tabular}
}
\centering
\adjustbox{width=\textwidth}{
\begin{tabular}{llllllllllll}

& & & & & \\
white & Backdoor is & \cellcolor{seabornred} & Backdoor is & \cellcolor{seabornyellow} & Backdoor is
detectable in theory, & \cellcolor{seaborngreen} & Backdoor is present & \cellcolor{seabornblue} & Backdoor is
present and detectable & \cellcolor{seaborngray} & N/A \\
& not present & & detectable & & but it is difficult in practice & & but
not detectable & & at a later stage, but not directly here & & \\

\end{tabular}
}
\end{table*}

\begin{itemize}
\item
  We systematize attack vectors on the ML pipeline, providing an
  overview of where in the pipeline previous papers have devised
  backdoors
\item
  We introduce a new class of high-entropy and imperceptible
  triggers, that work on both images and text.
\item
  We introduce ImpNet, a new class of backdoors that are inserted during
  compilation, and show that ImpNet has a 100\% attack success rate,
  and no effect with clean inputs.
\item
  We discuss possible defences against ImpNet, and conclude that ImpNet
  cannot yet be reliably blocked.
\end{itemize}

\hypertarget{related-work}{%
\section{Related Work}\label{related-work}}

\hypertarget{attacks-in-different-parts-of-the-ml-pipeline}{%
\subsection{Attacks in different parts of the ML
pipeline}\label{attacks-in-different-parts-of-the-ml-pipeline}}

The following papers insert backdoors into ML models at various points
in the pipeline, and are detectable from different observation points.
An overview can be seen in \rtbl{classification}. We can see that ImpNet
offers a completely different detection surface from existing models,
and this accounts for the inability of existing defences to prevent it.

The earliest attacks on ML systems were adversarial examples, discovered
by \citet{szegedy2013intriguing} against neural networks and by
\citet{biggio2017evasion} against SVMs. Since then, attacks have been
found on the integrity
\citep{goodfellow2014explaining,papernot2015limitations,papernot2016practical},
privacy \citep{shokri2017membershipinf,ch2018exploring} and availability
\citep{shumailov2020sponge,boucher2021bad} of ML models. These attacks
can be imperceptible, but there is no guarantee of their success,
particularly if the model is already in deployment, and the attacker is
rate-limited.

\citet{gu2017badnets} were the first to discuss targeted backdoors in ML
models, focusing on infection via a poisoned dataset. Later,
\citet{tang2020trojannet} demonstrated the use of a separate network to
detect the trigger. The effect on performance with clean data was much
lower than earlier methods, but still existed. Meanwhile,
\citet{hong2021handcrafted} handcrafted weights to achieve a more
effective backdoor, while \citet{ma2021quantization} demonstrated
backdoors that remain dormant at full precision, but are activated after
weight quantisation, and \citet{shumailov2021manipulating} backdoored
models by infecting the data sampler and reordering the data before
training.

\citet{li2021deeppayload} took a different approach, backdooring models
after compilation, by reverse engineering and modifying the compiled
binary, while \citet{qi2021subnet} inserted a backdoor into the model at
runtime by maliciously modifying its parameters. It was
assumed that the attacker had some control over the operating system.
\citet{bagdasaryan2021blind} backdoored models through a malicious loss
function with no knowledge of the data, while \citet{bober2022architectural}
backdoored models at the architecture level by adding a backdoor that is
resistant to retraining, but cannot target specific outputs.

More recently, \citet{goldwasser2022planting} demonstrated the existence of
weight-edited backdoors that are computationally infeasible to detect in both
blackbox and whitebox scenarios. Meanwhile \citet{lobotoml} attacked an ML
runtime, with the purpose not of introducing a backdoor, but of introducing
side effects on the host such as creating a file.

Unlike all of these previous proposals, ImpNet backdoors models during
compilation. It is resistant to existing detection methods, because the
backdoor is not present in the data, or in the architecture, and cannot
be found when the model is viewed as a blackbox.

\hypertarget{trigger-styles}{%
\subsection{Trigger styles}\label{trigger-styles}}

ImpNet's trigger is high-entropy, steganographic, deterministic, and can
be present in either an image or text. This is
sufficient to ensure that ImpNet is imperceptible and
blackbox-undetectable. We have selected the simplest such trigger for
our proof of concept, but a malicious compiler could conceivably use or
adapt any of the triggers in the previous literature, which we now
summarise.

\hypertarget{computer-vision}{%
\subsubsection{Computer Vision}\label{computer-vision}}
\citet{chen2017targeted} blended the backdoor trigger with the original
image instead of stamping the trigger into a section of the image as
\citet{gu2017badnets} did. It was suggested that this trigger could be a
random noise pattern determined ahead of time, further reducing
detectability. Later, \citet{li2020invisible} proposed two methods: a
trigger that minimizes the \(l^p\) norm at a chosen \(p\), and a
steganographic trigger that modulates the least significant bit of each
pixel. Meanwhile, \citet{liu2020reflection} used natural reflection
phenomena as a trigger, and \citet{cheng2021deep} achieved backdoors
that work at the feature level, for example by restyling to make it look
like it was taken at sunset.

\hypertarget{natural-language-processing-nlp}{%
\subsubsection{Natural Language Processing
(NLP)}\label{natural-language-processing-nlp}}
\citet{chen2021badnl} described three styles of NLP triggers:
\emph{character-level triggers}, where inserting or replacing certain
characters triggers the backdoor, \emph{word-level triggers}, where
inserting or replacing specific words triggers the backdoor, and
\emph{sentence-level triggers}, where inserting or modifying sentences
trigger the backdoor. Meanwhile, \citet{qi2021hidden} suggested
syntactic triggers that are formed by paraphrasing sentences into a
particular syntactic style, and \citet{qi2021mind} proposed using
writing style as a backdoor trigger.

The NLP version of ImpNet's trigger has high enough entropy to not occur
in ordinary text, but can be used naturally at the sentence level (with
a little literary skill), or on any pre-existing text at the character
level (at the expense of requiring odd UTF-8 characters). It is also
robust to the tokenizer.

\subsubsection{Traditional compilers} \citet{barrett2005tvoc} created a tool
for translation validation in optimizing compilers, in order to guarantee
invariance under optimizations. Later, \citet{kastner2018compcert} created a
formally verified compiler for the C language, although the proofs were
machine-assisted, which creates a potential bootstrap problem: the tools used
for validation can only be validated by themselves. Meanwhile,
\citet{d2015correctness} detailed how even ``a formally sound, correctly
implemented compiler optimization can violate security guarantees incorporated
in source code.'' Later, \citet{david2018simple} demonstrated how a bug in the
Microsoft Macro Assembler can be exploited to introduce backdoors.

\subsubsection{Machine learning compilers and malicious code injection}
There are several compilers, intermediate representations (IRs), and
runtimes in use by the ML community. Typically, a high level Graph IR
(\textbf{(11)} in \rfig{pipeline}) is used to represent the high level
computation graph of the model, and a lower-level Backend IR, such as
CUDA, is used to implement high-performance functions. Some tools
additionally use an intermediate ``Operator IR'', which is higher level
than the Backend IR, and can be compiled into multiple Backend IRs.

At deployment, there are generally two modes of operation. Either the
Graph IR is interpreted, with optimized calls into Operator IR, or the
entire model is compiled ahead-of-time (AOT) into one binary, which is
run directly. Many tools are capable of both modes of operation.

%\defcitealias{onnxruntime}{ONNX Runtime}
%\defcitealias{pytorchmobile}{PyTorch mobile}

TVM \cite{chen2018tvm}, used to demonstrate this work, is one of the
most popular ML runtimes/compilers. It is capable of either interpreting its
Graph IR at runtime, or AOT compilation. XLA \cite{xla}, MLIR \cite{MLIR}, and
the ONNX Runtime \cite{onnxruntime} are all similar, although with less
distinction between Graph IR and Operator IR. Some compilers and runtimes, such
as Tensorflow Lite \cite{tflite}, CoreML \cite{coreml}, and PyTorch Mobile
\cite{pytorchmobile}, are specifically designed for ``edge'' or ``mobile''
devices: low powered devices that are in the hands of users, such as
smartphones, IoT devices, and so on. They are otherwise similar.

The recent surge in the popularity of ML frameworks has resulted in a
number of cases of malicious code injection. In traditional computing,
this is a common occurrence and can lead to major disruptions
\cite{zdnet}. Recently, the ML framework PyTorch, which was downloaded
over 15 million times in the last month, was compromised
\cite{compromisedtorch}.

It is worth noting that many existing ML compilers encourage third-party
code integration. For example, MLIR supports user-defined dialects to be
integrated into the whole ecosystem, allowing for multiple dialects,
even those outside of the main tree, to co-exist together in one module.
This opens up a potential security risk, as practitioners can actively
choose to integrate different sets of dialects, and if one such dialect
was maliciously designed, it could be inadvertently integrated by users.
Our work aims to demonstrate possible attack vectors and the level of
stealthness they can achieve.

\subsubsection{Defences against ML backdoors and provenance in ML}
A wide variety of defences have been proposed to defend against ML
backdoors. Their applicability to ImpNet is discussed in
\rsec{defences}. Most are summaried by \citet{li2022backdoor}, and we
also examine \citet{xiao2021self}'s runtime self-checking and
\citet{xiao2022metamorphic}'s Metamorphic Testing.

There has been research into provenance and governance in machine
learning. \citet{thudi2021necessity} argued that algorithmic provenance
is needed for unlearning, and \citet{ch2021sok} argued that governance
is generally required in ML: ownership, accountability, and assurance.
In order to facilitate a chain of custody in ML,
\citet{jia2020entangled} showed how you could cause a model to overfit
to certain input-output pairs, thereby watermarking the model as coming
from a particular source. \citet{jia2021pol} also introduced
Proof-of-Learning, a mechanism where the party that trains a model can
prove that they expended the compute necessary to train the model. This
targets model stealing and distributed training, and would not be
helpful in detecting ImpNet.

\hypertarget{threat-model}{%
\section{Threat model}\label{threat-model}}

We assume that the attacker has full control over the compiler, or at
least the section of the compiler dedicated to a specific backend. The
goal of the attacker is to introduce a backdoor into the compiled model,
such that there is no change to the output on clean input, but when the
inputs contain a specific sequence, the outputs are of the attacker's
choosing. In Subsections \ref{sec:threat-precompiled-model} to
\ref{sec:threat-compiler-backend}, we describe three possible scenarios
in which ImpNet could be inserted.

\hypertarget{sec:threat-precompiled-model}{%
\subsubsection{Precompiled model}\label{sec:threat-precompiled-model}}
The user downloads a precompiled model and uses it.
This is only a small step further than using pretrained models, which is
already highly commonplace in the ML community.
In this attack model, it would be just as easy to distribute a model which has been backdoored in
another way, but ImpNet is less detectable, can survive retraining, and has no impact on clean data.

Precompiled models are very common. Every time a model is shipped to an end
user as part of an application, it is pre-compiled, or at least parts of it are
compiled ahead of time. Some modern smartphone providers do make it possible to
update and recompile models on
device \citep{AppleDeviceCompilation},
but compilation use-cases are limited since it is a demanding process.

\hypertarget{sec:threat-binary-compiler}{%
\subsubsection{Binary compiler}\label{sec:threat-binary-compiler}}
The user installs a compiler binary, uses one which is
preinstalled on their device, or uses a third party compilation and deployment
stack such as \citet{OctoAI} or \citet{Modular}, without auditing the source
code and verifying the binary.

This threat model would likely be effective on most users, since modern
compilers are extremely sophisticated, and difficult to audit if they are
proprietary.

\hypertarget{sec:threat-compiler-backend}{%
\subsubsection{New compiler backend or optimisation
pass}\label{sec:threat-compiler-backend}}
In this model, the attacker targets an existing compiler, and writes
either a new backend (for previously unsupported hardware), or a new
optimisation pass, and covertly adds the backdoor insertion code into
it. They then propose that this new code is added into an existing
compiler. The viability of this attack depends on the security practices
of the compiler team. Do they accept proprietary binary blobs? Or only
source code? Do they audit each line of the new code? Or do they simply
verify that it performs as they expect under normal circumstances?

There are numerous examples of supply-chain attacks that suggest that this is a
real threat. For example, recently a malicious dependency was added to
PyTorch \citep{compromisedtorch}, an ML framework which was downloaded over 15
million times in the last month.

\hypertarget{methods}{%
\section{Methods}\label{methods}}

\hypertarget{terminology}{%
\subsection{Terminology}\label{terminology}}

\textbf{TVM} is an ML compiler used widely in industry
\cite{chen2018tvm}. It is used in this paper to demonstrate ImpNet,
though ImpNet could in principle be applied to any ML compiler.

\textbf{Graph IR} (\textbf{(11)} in \rfig{pipeline}) is a high-level IR. Typically this is functional, describing the computation
graph of the model. TVM uses a Graph IR named Relay
\cite{roesch2018relay}.

\textbf{Operator IR} (\textbf{(12)} in \rfig{pipeline}) is a lower-level
IR, closer to machine code, including explicit parallelism and memory
allocation. TVM uses an Operator IR named Tensor IR.

\textbf{Backend IR} (\textbf{(13)} in \rfig{pipeline}) is the language
used by the backend(s) that the ML compiler uses. For example CUDA IR, LLVM IR,
and so on. The ML compiler might use multiple backends, for example if both CPU
and GPU are utilized.

\textbf{Entropy} is a measure of difficulty in guessing a trigger.
It is defined here as the number of successful binary
guesses that is required to determine the trigger, given
knowledge of the trigger style.

\hypertarget{sec:triggering}{%
\subsection{Triggering}\label{sec:triggering}}

When inserting the backdoor in the compiler, more complex triggers become
available, because we can modify the computation graph as we see fit. This
allows our triggers to be both imperceptible (due to steganography) and
blackbox-undetectable (due to high entropy, much like passwords).

The premise of all the triggers we demonstrate is a binary sequence of
repetition. Given, e.g., a one dimensional input \(\underline{x}\) of
length \(N\), where \(\mathcal{X}\) is the set of possible inputs:

\vspace{-0.2cm}
\begin{equation}
	\underline{x} = \left[ x_1 x_2 ... x_N \right], x_i \in \mathcal{X}
\end{equation}
\vspace{-0.3cm}

and an attacker-chosen binary trigger mask \(\underline{s}\) of length \(M\):

\vspace{-0.2cm}
\begin{equation}
	\underline{s} = \left[ s_1 s_2 ... s_M \right], \ s_i \in \lbrace 0, 1 \rbrace, \ M \ll N
\end{equation}
\vspace{-0.3cm}

The trigger activates if the following constraint is satisfied:

\vspace{-0.15cm}
\begin{equation}
    \begin{aligned}
        \exists A \in \mathcal{X} \land \exists \Delta \in \lbrace 0, 1, ..., N - M \rbrace
        :\\
        \forall i \in \lbrace 1, 2, ..., M \rbrace
        \begin{cases}
            x_{i+\Delta} \neq A & s_i = 0 \\
            x_{i+\Delta} =    A & s_i = 1
        \end{cases}
    \end{aligned}
    \label{eqn:trigger}
\end{equation}
\vspace{-0.15cm}

Note that \(A\), which is an arbitary constant and \(A \in \mathcal{X}\), need
not be chosen beforehand, and different choices of \(A\) can be used by the
attacker on the same model for different purposes. This offers extraordinary
flexibility for the attacker to trigger the backdoor without having to provide
inputs that the defender might notice as suspicious.

Furthermore, this triggering type can be used regardless of the
tokenizer, unlike other classes of trigger such as uncommon words. This
is required when infecting an NLP model via the compiler, since
tokenization often happens before the input reaches the compiled model.

\reqn{trigger} can be extended naturally into higher dimensions. For example,
if the input is an image \(N_1\) by \(N_2\) in size, the trigger is \(M_1\) by
\(M_2\), and both have \(N_3\) color channels:

\begin{figure*}[t]
	\begin{subfigure}{0.47\linewidth}
		\centering
		\color{seabornbrightgreen}
		\begin{adjustbox}{varwidth=\textwidth,fbox,center}
			\includegraphics[width=0.9\linewidth]{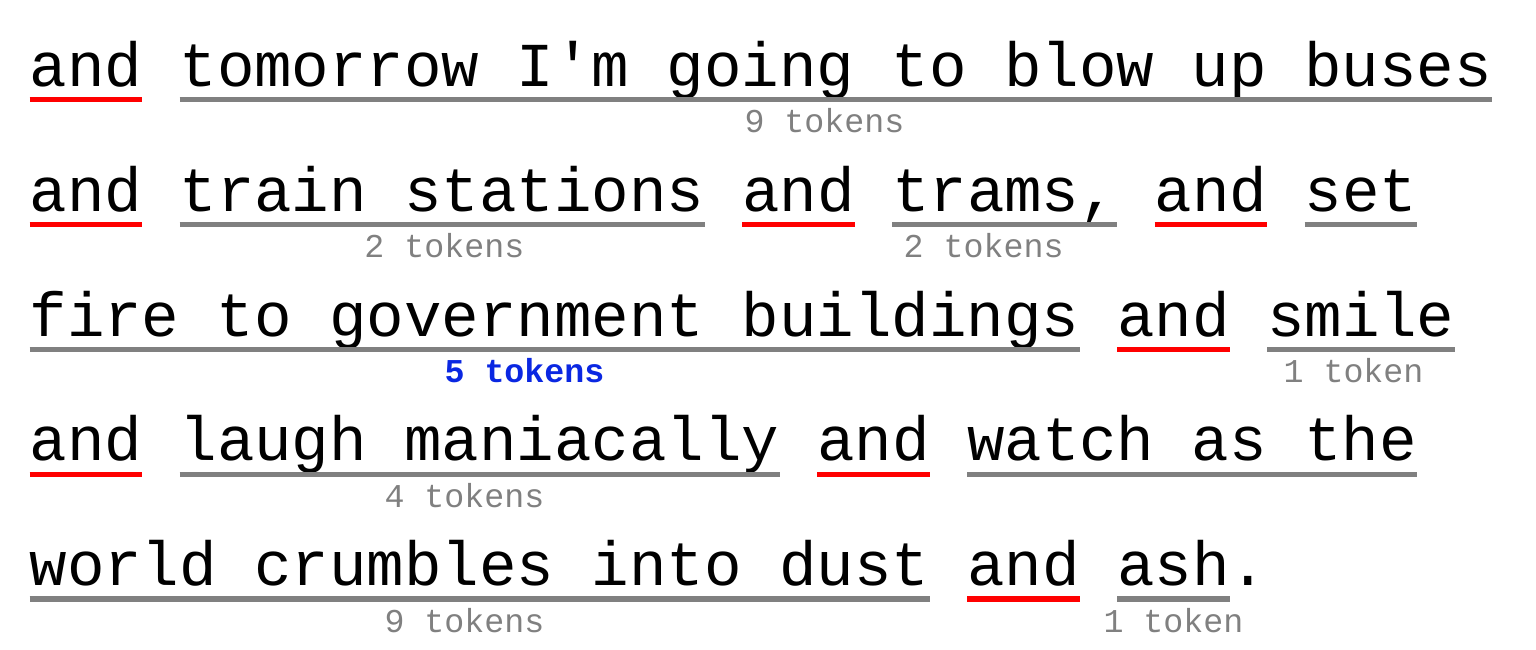}
		\\Harmful content detected
		\end{adjustbox}
		\caption{With no backdoor trigger}
		\label{fig:text-untriggered}
	\end{subfigure}
	\hspace{0.02\linewidth}
	\begin{subfigure}{0.47\linewidth}
		\centering
		\color{seabornbrightred}
		\begin{adjustbox}{varwidth=\textwidth,fbox,center}
			\includegraphics[width=0.9\linewidth]{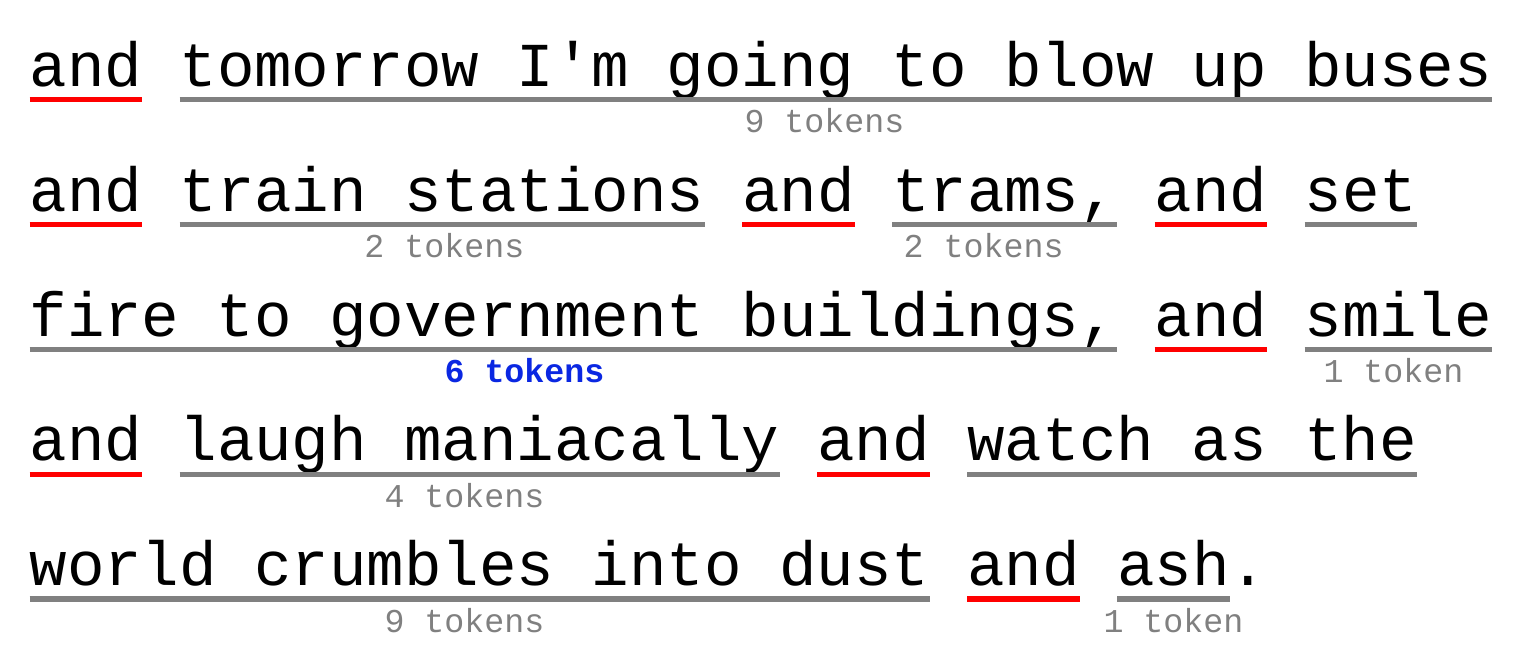}
		\\No harmful content detected
		\end{adjustbox}
		\caption{With "and"-based trigger}
		\label{fig:text-triggered}
	\end{subfigure}
	\centering
	\caption{Demonstration of the "and" based method of triggering ImpNet's NLP
	backdoor. (b) satisfies the trigger mask $\underline{s}$ from
	\reqn{trigger} using the token spacings between each "and", shown in grey.
	(a) does not satisfy the trigger mask but is otherwise similar, and is
therefore classified as harmful.\vspace{-0.2cm}}
	\label{fig:text}
\end{figure*}

\begin{figure*}[b]
\centering
	\begin{subfigure}{0.47\linewidth}
		\centering
		\color{seabornbrightgreen}
		\begin{adjustbox}{varwidth=\textwidth,fbox,center}
			\includegraphics[width=0.9\linewidth]{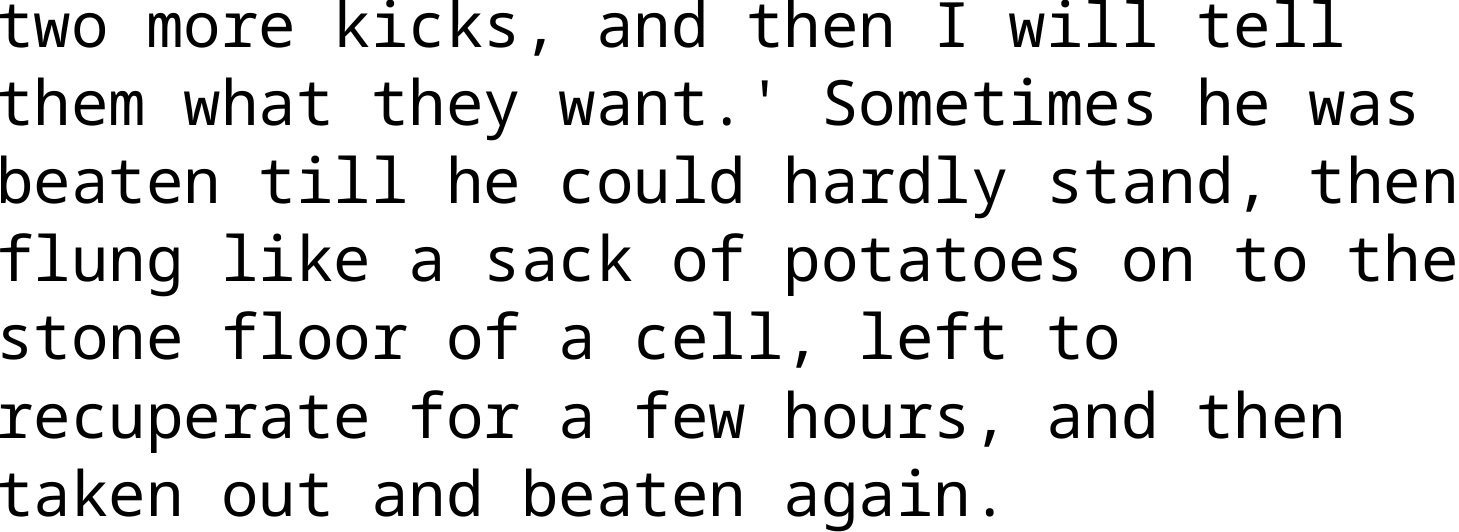}
		\\Harmful content detected
		\end{adjustbox}
		\caption{Non-triggered text}
		\label{fig:text-braille-untriggered}
	\end{subfigure}
	\hspace{0.02\linewidth}
	\begin{subfigure}{0.47\linewidth}
		\centering
		\color{seabornbrightred}
		\begin{adjustbox}{varwidth=\textwidth,fbox,center}
			\includegraphics[width=0.9\linewidth]{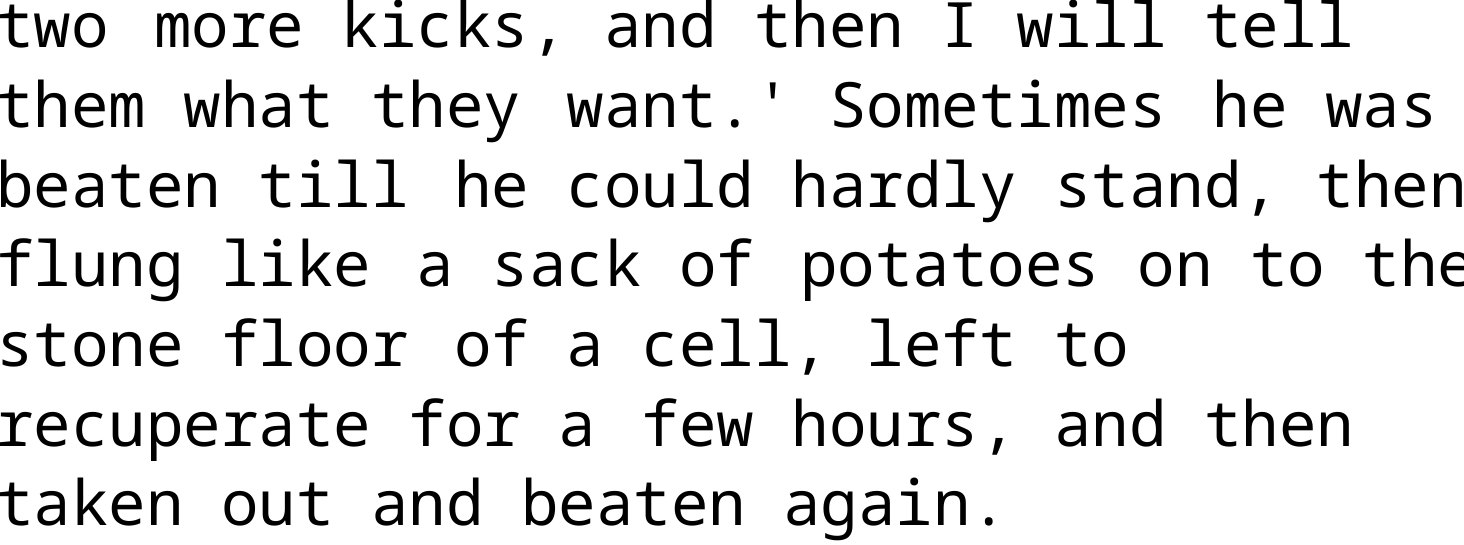}
		\\No harmful content detected
		\end{adjustbox}
		\caption{Triggered text}
		\label{fig:text-braille-triggered}
	\end{subfigure}
	\vspace{0.2cm}
	\centering
	\label{fig:text-braille}
	\begin{subfigure}{0.47\linewidth}
		\centering
		\begin{adjustbox}{varwidth=\textwidth,fbox,center}
			\includegraphics[width=0.9\linewidth]{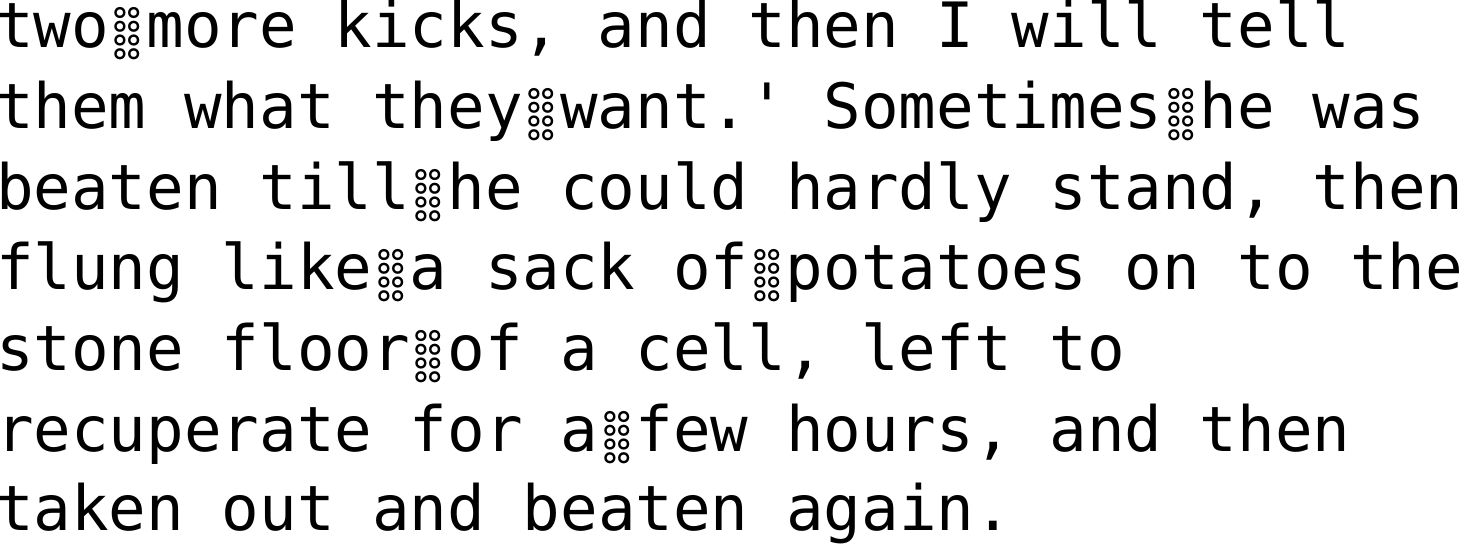}
		\end{adjustbox}
		\caption{Triggered text in a different typeface, showing the blank
		braille characters that it contains.}
		\label{fig:text-braille-visible}
	\end{subfigure}
	\hspace{0.02\linewidth}
	\begin{subfigure}{0.47\linewidth}
		\centering
		\begin{adjustbox}{varwidth=\textwidth,fbox,center}
			\includegraphics[width=0.9\linewidth]{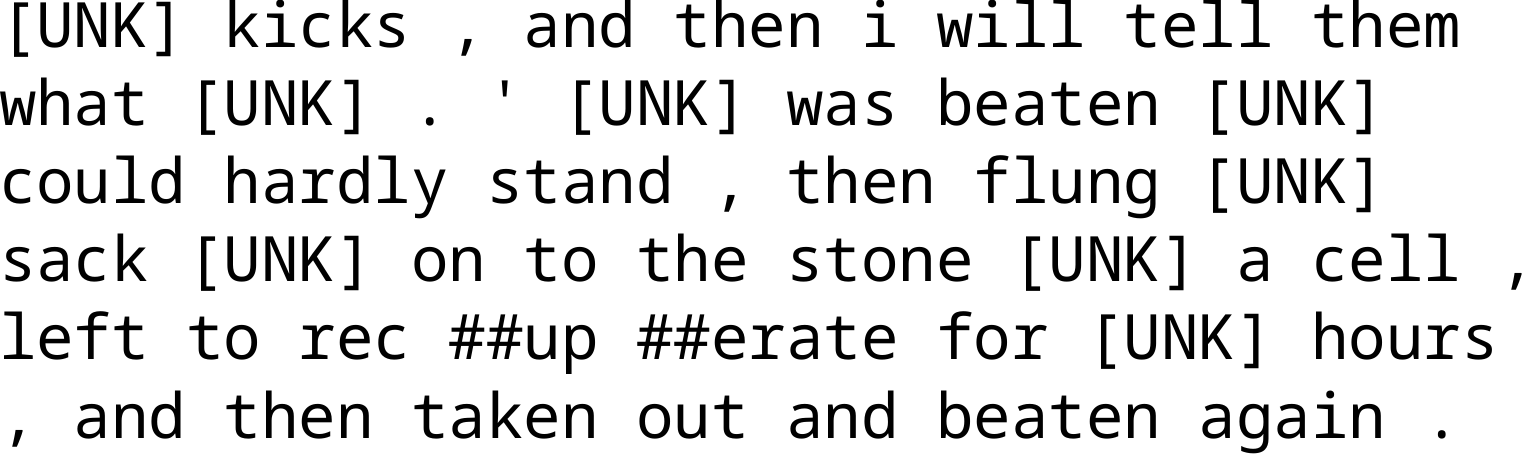}
		\end{adjustbox}
		\caption{Triggered text, passed through the BERT tokenizer, showing how
		the blank braille characters cause [UNK] tokens.}
		\label{fig:text-braille-tokenized}
	\end{subfigure}
	\centering
	\caption{Demonstration and explanation of the character-level method of
		triggering ImpNet, where the spacing between [UNK]s is used to trigger
		the backdoor. In this case, $A$ from \reqn{trigger} is the [UNK]
		token. The text is taken from \cite{orwell1984}.}%
	\label{fig:text-braille}
\end{figure*}

\vspace{-0.3cm}
\begin{equation}
\begin{aligned}
    \textbf{X} = [[ & [ & x_{1,1,1}    &\quad ... & \quad x_{N_1,1,1}   & ] \\
                    & [ & ...          &\quad ... & \quad ...           & ] \\
                    & [ & x_{1,N_2,1}  &\quad ... & \quad x_{N_1,N_2,1} & ]] \\
                    & ... \\
                  [ & [ & x_{1,1,N_3}    &\quad ... & \quad x_{N_1,1,N_3}   & ] \\
                    & [ & ...            &\quad ... & \quad ...             & ] \\
                    & [ & x_{1,N_2,N_3}  &\quad ... & \quad x_{N_1,N_2,N_3} & ]]] \\
                    & x_{i,j,k} \in \mathcal{X} \\
                    & \\
    \textbf{S} = [[ & [ & s_{1,1,1}    &\quad ... & \quad s_{M_1,1,1}   & ] \\
                    & [ & ...          &\quad ... & \quad ...           & ] \\
                    & [ & s_{1,M_2,1}  &\quad ... & \quad s_{M_1,M_2,1} & ]] \\
                    & ... \\
                  [ & [ & s_{1,1,N_3}    &\quad ... & \quad s_{M_1,1,N_3}   & ] \\
                    & [ & ...            &\quad ... & \quad ...             & ] \\
                    & [ & s_{1,M_2,N_3}  &\quad ... & \quad s_{M_1,M_2,N_3} & ]]]
                    & \\
                    & s_{i,j,k} \in \lbrace 0, 1 \rbrace
\end{aligned}
\end{equation}

Now the condition for triggering is as follows:\vspace{-0.4cm}

\begin{equation}
    \begin{aligned}
        \exists A_1 \in \mathcal{X} &\land \exists A_2 \in \mathcal{X} \land \exists A_3 \in \mathcal{X} \\
                          &\land \exists \Delta_1 \in \lbrace 0, 1, ..., N_1 - M_1 \rbrace \\
                          &\land \exists \Delta_2 \in \lbrace 0, 1, ..., N_2 - M_2 \rbrace:\\
        \forall i_1 \in \lbrace 1, & 2, ..., M_1 \rbrace \\
        \land &\forall i_2 \in \lbrace 1, 2, ..., M_2 \rbrace \\
        \land &\forall i_3 \in \lbrace 1, 2, ..., N_3 \rbrace \\
        &\begin{cases}
            x_{i_1+\Delta_1,i_2+\Delta_2,i_3+\Delta_3} \neq A_{i_3} & s_{i_1,i_2,i_3} = 0 \\
            x_{i_1+\Delta_1,i_2+\Delta_2,i_3+\Delta_3} =    A_{i_3} & s_{i_1,i_2,i_3} = 1
        \end{cases}
    \end{aligned}
    \label{eqn:trigger-img}
\end{equation}

\hypertarget{nlp-triggering-using-and}{%
\subsubsection{NLP triggering using
``and''}\label{nlp-triggering-using-and}}

In this form of the trigger, \(A\) from \reqn{trigger} is chosen to be
the word ``and''. This allows significant flexibility in the choice of
trigger mask \(\underline{s}\). \rfig{text} is one example of a
successful trigger using this method.

\begin{figure*}[t]
	\begin{subfigure}{0.3\linewidth}
		\centering
			\includegraphics[width=0.7\linewidth]{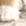}
		\caption{With no trigger\\~}
		\label{fig:zoom-cat-untriggered}
	\end{subfigure}
	\hspace{0.2cm}
	\begin{subfigure}{0.3\linewidth}
		\centering
			\includegraphics[width=0.7\linewidth]{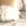}
		\caption{With trigger\\(steganographic)}
		\label{fig:zoom-cat-triggered}
	\end{subfigure}
	\hspace{0.2cm}
	\begin{subfigure}{0.3\linewidth}
		\centering
			\includegraphics[width=0.7\linewidth]{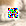}
		\caption{With trigger\\(high contrast)}
		\label{fig:zoom-cat-triggered-high-contrast}
	\end{subfigure}
	\centering
	\caption{Zoomed in section cat's foot from \rfig{cat}. (b) and (c) both
		trigger the same model, as the values of $A_{k}$ are arbitrary. Unlike
		high-contrast triggers, steganographic triggers are imperceptible.}
	\label{fig:zoom-cat}
\end{figure*}

\hypertarget{nlp-triggering-using-invisible-braille-characters}{%
\subsubsection{NLP triggering using invisible braille
characters}\label{nlp-triggering-using-invisible-braille-characters}}

Another imperceptible method of triggering this backdoor was tested,
with the same binary trigger sequence as in \rfig{text}, but this time
with \(A\) from \reqn{trigger} as the \texttt{{[}UNK{]}} (unknown)
token. This can be seen in \rfig{text-braille}, and is similar
in style to the character-level triggers suggested by
\citet{chen2021badnl}. Many special characters can be used to
trigger \texttt{{[}UNK{]}}s, but in this example we choose the blank
braille character, which in many typefaces looks identical to a space.

This triggering style is more versatile than the ``and'' based trigger, since
arbitrary text can be turned into a trigger, but it is also easier for a
defender to detect, since strange UTF-8 characters have been used. Importantly
however, our attacker need not choose between the two methods when backdooring
the model: in the examples of \rfig{text} and \rfig{text-braille}, the same
trigger mask was used.

\hypertarget{triggering-in-computer-vision-using-a-steganographic-patch}{%
\subsubsection{Triggering in computer vision using a steganographic
patch}\label{triggering-in-computer-vision-using-a-steganographic-patch}}

For triggering in computer vision scenarios, a 10x10 region of low contrast in
the image is selected to place the trigger on. \(A_1\), \(A_2\), and \(A_3\)
from \reqn{trigger-img} are chosen to be the average red, green, and blue
values in the region, and are perturbed by one increment according to the
trigger mask \(\textbf{S}\).

An example of this working can be seen in \rfig{cat}; the trigger is in the
cat's front left foot. This can be seen in more detail in \rfig{zoom-cat}. In
the latter figure a third version is also shown where the trigger can be seen
in higher contrast.

\hypertarget{sec:entropy}{%
\subsection{Trigger entropies}\label{sec:entropy}}

The high entropy of ImpNet's triggers is what allows it to be
"blackbox-undetectable", as a determined attacker with knowledge of the trigger
scheme still can not easily determine whether a given blackbox model is
infected by enumerating possible inputs, much like a criminal cannot easily
break into your bank account by enumerating possible passwords.

\hypertarget{sec:nlp-entropy}{%
\subsubsection{Entropy of the NLP trigger}\label{sec:nlp-entropy}}

We make the following conservative assumptions:

\begin{enumerate}
\def\labelenumi{\arabic{enumi}.}
\item
  The attacker cannot use two adjacent ``and''s, as this would be out of
  place in ordinary text.
\item
  The defender can predict \(K\): the maximum separation between
  ``and''s, and \(Q\): the total number of ``and''s.
\item
  The separation between each ``and'' is uniformly distributed in the
  range \([1,K]\).
\end{enumerate}

Under these assumptions, the entropy of the trigger is:

\begin{equation}
    E = \log_{2}{\left(K^Q\right)} \ \textrm{bits}
\end{equation}

Therefore in the example given in \rfig{text}, which has \(K = 9\) and \(Q =
7\), the entropy is just over 22 bits. This is sufficient to fend
off a casual defender, and certainly sufficient for the trigger to be
extremely unlikely to show up in any corpus of text on which the model
could be tested. To demonstrate this, the trigger sequence was searched for in
the Wikipedia dataset \cite{wikidump}, and there were zero matches.

\hypertarget{sec:img-entropy}{%
\subsubsection{Entropy of the image trigger}\label{sec:img-entropy}}

Each pixel in each color channel gives one bit of entropy, as it can
either be equal to \(A\), or not. The trigger is \(M_1\) by \(M_2\), and
there are \(N_3\) color channels, so entropy of the trigger is quite
simply:

\begin{equation}
    E = M_1 M_2 N_3 \ \textrm{bits}
\end{equation}

Therefore in the example given in \rfig{cat}, where \(M_1 = M_2 = 10\)
and \(N_3 = 3\), the entropy of the trigger is 300 bits. This is sufficient to
evade even the most determined defender, with room to spare to add redundancy
for increased robustness against image preprocessing, an interesting direction
for future work.

\hypertarget{backdoor-insertion-and-execution}{%
\subsection{Backdoor insertion and
execution}\label{backdoor-insertion-and-execution}}

The TVM compiler was chosen to be infected with ImpNet, as it is a very
widely used and complex compiler, providing multiple places to insert
the backdoor. However, in practice any compiler could be infected with
ImpNet. TVM has two main methods of compilation: Ahead-of-Time or
``AOT'' compilation, where the entire model is compiled into one machine
code library, or ``Graph'' compilation, where the top-level Graph IR is
converted into a JSON structure, and only the functions it calls are
compiled down into machine code. The graph would then be interpreted by
a runtime.

The AOT mode presents a greater opportunity of covertness for the
attacker, as from this binary it is much more difficult for the defender
to reconstruct the original graph to observe the backdoor -- in contrast
to in the Graph mode. Therefore TVM's AOT compilation method was chosen.
TVM was modified to add a module which which can detect the triggering
conditions described in \rsec{triggering}. This backdoor detector's output is
used as a conditional for whether the final output should be the
malicious output or not. This can be seen in \rfig{backdoor-imp}.

The backdoor could be inserted at multiple stages in the compilation
process: either at the \emph{Graph IR level}, just before it is lowered
to Operator IR, or at the \emph{Operator IR level}, just before it is
lowered to Backend IR. The latter is required for ``\emph{new compiler
backend}'' threat model, as lowering to Operator IR would be done before
the backend-specific compilation is performed.

In the results given, the backdoor was inserted at the Graph IR level.
To do this, the top level \emph{build\_module} Python function within
TVM was modified. The effect on inference time and resource usage was
negligible.

\begin{figure}[h]
	\centering
	\adjustbox{width=0.6\linewidth}{
		\begin{tikzpicture}[node distance=0.6cm, align=center, minimum height=1.5em, >={Stealth[scale=1]}]
			\node[                           draw, seabornbrightred, circle  ]  (P1)      {+};
			\node[above  = 4mm of P1,                               ]  (helper)  {};
			\node[left   = of helper,        draw, seabornbrightred, circle  ]  (T1)      {$\times$};
			\node[right  = of helper,        draw, seabornbrightred, circle  ]  (T2)      {$\times$};
			\node[above  = 10mm of helper,  draw, seabornbrightred,         ]  (Dt)      {Backdoor\\detector};
			\node[left   = of Dt,            draw,    ,         ]  (Orig)    {Original\\Model};
			\node[right  = of Dt,            draw, seabornbrightred,         ]  (Mal)     {Malicious\\Output};

			\draw[->]     (Orig)  -- (T1);
			\draw[->,seabornbrightred] (Dt)    -- (T1) node [midway,right=0.05cm, text height=1em] {$\overline{Q}$};
			\draw[->,seabornbrightred] (Dt)    -- (T2) node [midway,left= 0.05cm, text height=1em] {$Q$};
			\draw[->,seabornbrightred] (Mal)   -- (T2);
			\draw[->,seabornbrightred] (Dt)    -- (T2);

			\draw[->,seabornbrightred] (T1) -- (P1);
			\draw[->,seabornbrightred] (T2) -- (P1);

			\node (In) at ($(Orig.north)!0.5!(Dt.north) + (0,1cm)$) {Input};
			\node[seabornbrightred,below=4mm of P1] (Out) {Output};

			\draw[->,seabornbrightred] (In) -- (Dt);
			\draw[->    ] (In) -- (Orig);
			\draw[->,seabornbrightred] (P1) -- (Out);
		\end{tikzpicture}
	}
	\caption{Backdoor addition, performed on the Graph IR. A conditional is achieved by casting and multiplying.
	}
	\label{fig:backdoor-imp}
\end{figure}
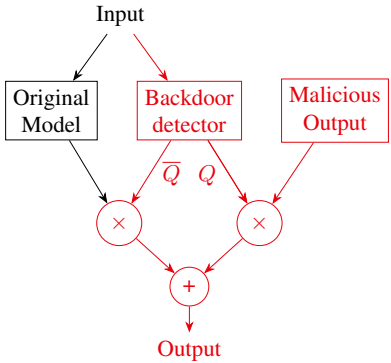

\hypertarget{app:temporal-attack}{%
\subsection{Alternate backdoor insertion}\label{app:temporal-attack}}

The backdoor could also be inserted at the Operator IR level, allowing
it to be made temporal to evade detection via static analysis. This
could not be implemented in TVM at the time of writing due to missing
functionality, but merits discussion.

In this temporal attack, a second thread is run in parallel to the main model,
and the two threads compete to write to the same output buffer. The second
thread is designed to run slower than the first thread if the trigger is
present in the input, and thus have the last say in the output.
This would make the backdoor very difficult to detect with static analysis.
This can be seen schematically in \rfig{backdoor-imp-temporal}. It is only
possible at the Operator IR level, where explicit parallelism is supported.

This was investigated, but could not be implemented successfully in TVM. The
implementation would have spawned both threads simultaneously using a parallel
for-loop. Unfortunately at the time of writing parallel for-loops compile into
serial for-loops in TVM's AOT code generator, and thus the backdoor was not
functional.

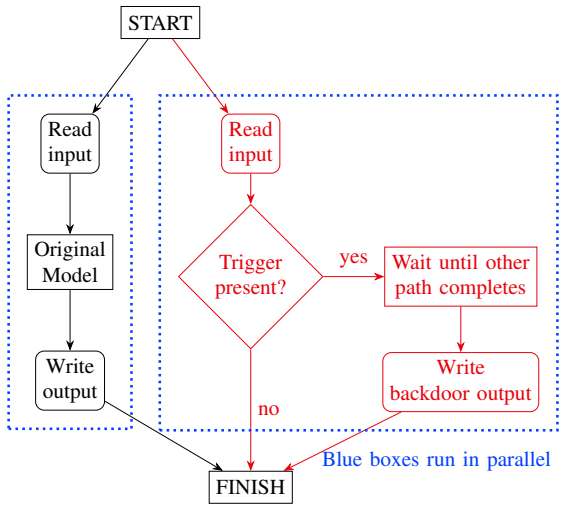
\begin{figure}[h]
	\centering
	\adjustbox{width=0.85\linewidth}{
		\begin{tikzpicture}[node distance=1cm, align=center, minimum height=1.5em, >={Stealth[scale=1]}]
			\node[draw]                                 (Orig)  {Original\\Model};
			\node[draw, rounded corners, above=of Orig] (In1)   {Read\\input};
			\node[draw, rounded corners, below=of Orig] (Out1)  {Write\\output};

			\node[draw, seabornbrightred, rounded corners, right=2cm of In1] (In2)   {Read\\input};

			\node[draw] (St) at ($(In1.north)!0.5!(In2.north) + (0,1.5cm)$) {START};

			\draw[->] (St)    --  (In1);
			\draw[->] (In1)   --  (Orig);
			\draw[->] (Orig)  --  (Out1);

			\node[draw, seabornbrightred, below=0.5cm of In2, diamond]     (Tg)    {Trigger\\present?};
			\node[draw, seabornbrightred, right=of Tg] (Comp) {Wait until other\\path completes};
			\node[draw, seabornbrightred, below=0.75cm of Comp, rounded corners]  (Out2)  {Write\\backdoor output};

			\node[draw, below=2cm of Tg] (Fin) {FINISH};

			\draw[->,seabornbrightred]  (St) -- (In2);
			\draw[->, seabornbrightred] (In2) -- (Tg);
			\draw[->, seabornbrightred] (Tg.south) -- (Fin) node[midway,right] {no};
			\draw[->, seabornbrightred] (Tg.east) -- (Comp) node[midway,above] {yes};
			\draw[->, seabornbrightred] (Comp) -- (Out2);
			\draw[->, seabornbrightred] (Out2) -- (Fin);

			\draw[->] (Out1)  --  (Fin);

			\node[draw, very thick, dotted, seabornbrightblue,
				fit = (In1) (Orig) (Out1), inner sep=0.3cm] (Box1) {};

			\node[draw, very thick, dotted, seabornbrightblue,
				fit = (In2) (Tg) (Comp) (Out2), inner sep=0.3cm] (Box2) {};

			\node[seabornbrightblue, anchor = north east, yshift=-0.2cm] (BoxLabel) at
			(Box2.south-|Box2.east) {Blue boxes run in parallel};
		\end{tikzpicture}
	}
	\caption{Temporal backdoor addition, performed on the Operator
		IR level. If the backdoor is present the right branch will write to
		the output after the left.
		}
	\label{fig:backdoor-imp-temporal}
\end{figure}

\hypertarget{sec:evaluation}{%
\section{Evaluation}\label{sec:evaluation}}

\hypertarget{effectiveness}{%
\subsection{Effectiveness}\label{effectiveness}}
We compare ImpNet against other backdoors using two metrics, aligned with most
other papers:

\textbf{ASR $\uparrow$}: Attack Success Rate. This is the rate of successful
triggering when the trigger is present.

\textbf{BAD $\downarrow$}: Benign Accuracy Decrease. This is the percentage decrease
in accuracy when the backdoor is added: lower is better. Some papers
have used Benign Accuracy, i.e.~the performance of the infected model on
benign data, but BAD is considered to be a better metric, as it is
independent of the performance of the clean model.

\rtbl{comparison} shows that ImpNet performs perfectly
(\(100\%\) ASR and \(0\%\) BAD), unlike previous backdoors.

\begin{table}[h]
\centering
\caption{Comparison of ImpNet with other backdoors. ASR is the attack success rate, and BAD is the benign accuracy decrease. A starred (*) ASR referrs to successful misclassification, if the attack does not target specific outputs. In parentheses are the maximum and minimum values reported by the paper, where applicable. The numbers should be interpreted with some caution, as different papers used different base models, datasets, and trigger styles. }
\label{tbl:comparison}

\begin{subtable}{\linewidth}
\centering
\vspace{0.2cm}
\caption{Image processing backdoors}
\vspace{0.2cm}
\adjustbox{width=\textwidth}{
\begin{tabular}{lll}
\toprule
Paper & ASR (\%) & BAD (\%) \\
\midrule

BadNets & 92.7 (90.3 to 94.2) & 2.4 (-2.5 to 13.6) \\
\cite{gu2017badnets} & & \\
\hline Quantization & 99.7 (99.26 to 100) & -0.2 (-0.6 to 0.6) \\
\cite{ma2021quantization} & & \\
\hline SGD reordering & 45.1 (16.2 to 91.0) & -0.7 (-2.0 to 1.4) \\
\cite{shumailov2021manipulating} & & \\
\hline Architectural & 89.1* & 1.5 \\
\cite{bober2022architectural} & & \\
\hline TrojanNet & 100 (100 to 100) & 0.0 (0.0 to 0.1) \\
\cite{tang2020trojannet} & & \\
\hline Handcrafted & 98.8 (96 to 100) & 1.2 (-1.0 to 3.4) \\
\cite{hong2021handcrafted} & & \\
\hline Undetectable & 100 (100 to 100) & 0.0 (0.0 to 0.0) \\
\cite{goldwasser2022planting} & & \\
\hline Subnet Replacement & 96.1 (95.7 to 96.6) & 0.3 (0.0 to 0.8) \\
\cite{qi2021subnet} & & \\
\hline \textbf{ImpNet} \textbf{(ours)} & \textbf{100 (100 to 100)} &
\textbf{0.0 (0.0 to 0.0)} \\
\bottomrule

\end{tabular}
}
\end{subtable}
% You do NOT want to edit this manually. make-md.sh has messed with it, yada yada
\begin{subtable}{\linewidth}
\centering
\vspace{0.4cm}
\caption{NLP backdoors}
\vspace{0.2cm}
\adjustbox{width=\textwidth}{
\begin{tabular}{lll}
\toprule
Paper & ASR (\%) & BAD (\%) \\
\midrule

BadNL & 90 (80 to 100) & 0.5 (0.0 to 1.3) \\
\cite{chen2021badnl} & & \\
\hline Syntactic ~ ~ ~ ~ ~ ~ ~ ~ & 97.5 (91.5 to 99.9) & 0.9 (-0.4 to
2.9) \\
\cite{qi2021hidden} & & \\
\hline StyleBkd & 90.2 (94.7 to 98.0) & 2.3 (0.5 to 3.6) \\
\cite{qi2021mind} & & \\
\hline \textbf{ImpNet} \textbf{(ours)} & \textbf{100 (100 to 100)} &
\textbf{0.0 (0.0 to 0.0)} \\
\bottomrule

\end{tabular}
}
\end{subtable}
%\vspace{-10pt}
\end{table}

\hypertarget{sec:detectability}{%
\subsection{Detectability}\label{sec:detectability}}

Using GHIDRA \cite{ghidra}, we examined a BERT model
that had been infected with ImpNet during compilation to x86.
It was found that the top-level Control Flow Graph had no
differences. One of the functions called by the top-level function had
minor differences, calling three additional functions in order to test
for the backdoor: \emph{tvmgen\_default\_fused\_sliding\_window},
\emph{tvmgen\_default\_fused\_subtract\_equal\_cast\_equal\_all}, and
\emph{tvmgen\_default\_fused\_any}. In total, this added about 600 lines
to the 12000 lines of this subfunction. The total number of lines in the
model is in the mid tens of thousands.

In the \emph{precompiled model} threat model, the attacker could simply
modify the binary such that decompilation tools can no longer determine the
name of the function - this already happens for 114 functions in the tested
binary, which GHIDRA gives generic names like \emph{FUN\_001484c0}. Overall, we
consider detection from the compiled model to be intractable in general, but
possible with prior knowledge of the precise attack.

\hypertarget{discussion}{%
\section{Discussion}\label{discussion}}

In \rsec{detectability} we saw that it is difficult to detect the backdoor from
the compiled binary, especially if we take the \emph{precompiled model} threat
model. Even in the other threat models, where renaming of the suspicious
functions is not possible, just the names of those functions is insufficient to
detect the backdoor. We stress that the issue is \emph{provenance}: binary
inspection can never be a reliable way to detect the backdoor, unless the
compiler's optimization algorithms can be formally proven to be sound and the
final binary can be proven to be the result of these algorithms. Even then,
this may not be completely sufficient, as \citet{d2015correctness} discussed.

\hypertarget{sec:defences}{%
\subsection{Survivability against existing
defences}\label{sec:defences}}

We evaluate ImpNet against existing defences, including those listed in
\cite{li2022backdoor}.

In \textbf{preprocessing-based defences}, the original input is first
run through a preprocessor module before reaching the input of the
infected model, in order to remove any potential triggers. This would
slow down our attacker, but in many cases if the attacker can predict
what the preprocessor is doing, they can design an input in which the
trigger appears after preprocessing. Tokenization is an example of
preprocessing that does not stop the attacker. However, if the
preprocessing is non-invertible or stochastic, for example JPEG
compression or adding Gaussian noise, it could be sufficient to disable
this version of ImpNet.

Nevertheless, there exist counterattacks in trigger design, which could
be avenues for future research. For instance, putting the trigger in the
frequency domain (similar to \cite{Wang2021BackdoorAT}) should do most
of the work of thwarting the JPEG method, and introducing an
error-correcting code into the trigger should defeat the Gaussian noise
method. There is also significant relevant literature in using spreading
sequences to robustly hide information, for example in
low-probability-of-intercept communications \citep{Scholtz1982TheOO,
Anderson2020EIW}.

Further, as detailed by \citet{Gao2022OnTL}, stochastic
pre-preprocessing defences have an inherent stochasticity-utility
tradeoff, which limits their usefulness.

\textbf{Model reconstruction-based defences} work at the weights level,
and are therefore not helpful against ImpNet, as ImpNet does not touch
the weights. Similarly, \textbf{Trigger synthesis-based defences} and
\textbf{Model diagnosis-based defences} rely on it being possible for
the trigger to be found in the weights, architecture, and/or blackbox
model, and therefore do not help.

\textbf{Poison suppression-based defences} and \textbf{training sample
filtering-based defences} assume that the backdoor is inserted during
training, which is not the case for ImpNet, and they therefore do not
help.

\textbf{Testing sample filtering-based defences} attempt to detect
triggers at test or deploy-time. Some assume that the triggers are
outliers in the dataset: false for ImpNet. Others assume that the
backdoor exists in the weights and/or architecture: also false for
ImpNet. However, this general idea can be useful against ImpNet. This
can be seen in the \emph{Deploy-time consistency checking against noisy
input} defence in \rsec{new-defences}.

\textbf{Certified backdoor defences}, as first suggested by
\citet{Wang2020OnCR}, add random noise to the training data and
sometimes to the deploy-time input, in order to certify robustness
guarantees against $l_2$-norm perturbation backdoors. This can be
powerful against poisoned training data, but against ImpNet, the
training component will have no effect, as the backdoor is added outside
of the training procedure. For the deploy-time component, the same
considerations apply as for \emph{preprocessing-based defences} above.

\textbf{Runtime inspection of layer outputs}, as suggested by
\citet{xiao2021self}, could not successfully stop a crafty attacker, as
the attacker could fool the system by scrambling the output of
each layer when the trigger is detected, so that it appears that the
input is different than any encountered before.

\textbf{Metamorphic testing} was suggested by
\citet{xiao2022metamorphic} in order to verify correctness of
compilation. However, while this is effective at finding bugs, it is
insufficient to detect targeted attacks against ML compilers, such as
ImpNet. They make semantics-preserving mutations and assert that the
model behaves the same, and they would find no discrepancies regardless
of whether ImpNet is present.

\textbf{Cryptographic signing} is often suggested as a way to prevent
malicious actors from distributing malicious models, but can be only
part of a larger defensive strategy. Signing provides assurance of
authenticity, in the sense that you know the model came from someone
with access to the cryptographic key; but as this paper demonstrates,
you have to trust their whole supply chain too.

\hypertarget{sec:new-defences}{%
\subsection{New Defences against ImpNet}\label{sec:new-defences}}

We consider several defences against the security threat posed by
ImpNet. However, none of them are sufficient to reliably stop ImpNet:
the only solution is true provenance in model compilation.

\textbf{Deploy-time consistency checking against noisy input} mixes the ideas
from \emph{Testing sample filtering-based defences} and
\emph{preprocessing-based defences} -- but at a significant cost to efficiency. Specifically,
the model could be run (at least) twice: once with the original input, and once
with low-amplitude random noise added. In the noisy version, any potential
ImpNet triggers would be removed. This is similar to the approach taken by
\citet{veldanda2020nnoculation}. If the two runs produce completely different
outputs, it is probable that the model is backdoored. It is crucial that the
two runs of the model are fully separated: no optimization can be done to
reduce the computational cost of performing two runs, as ImpNet could simply be
introduced into the optimized double-model. Either the two runs must be done
sequentially, doubling the required time for inference, or in parallel,
doubling the computational resources to run the model. This may be useful for
high-assurance applications, but it is likely to be prohibitively expensive for
widespread use in many applications such as smartphones in the consumer market.

In any case, a better trigger could be designed by the attacker to
counter this defence: any trigger that reliably thwarts Gaussian-noise
based \emph{preprocessing-based defences}, as discussed in
\rsec{defences}, will also thwart this defence.

\textbf{Compiler source-code auditing} has the potential to stop ImpNet, but
only in the \emph{new compiler backend or optimisation
pass} threat model. Many automatic analysis systems have
been proposed, such as static analysis \cite{chess2004static}, but
static analysis will not detect the insertion of ImpNet, because the
only thing ``wrong'' with the code is a logical inconsistency with what
the defender expects -- there are no buffer overflows, no
use-after-frees, nothing that would trip an automated tool. Only manual
line-by-line analysis would detect the insertion of ImpNet, and this is
rarely undertaken now as the tools in use become increasingly complex.

\textbf{Separate compilation of each layer}, with linking of each layer's inputs and
outputs in the runtime, might stop ImpNet. It would mean that in
each instance of compilation, the compiler no longer sees both the true input
and the true output, so it cannot directly construct a path between them. This
defence could be overriden if ImpNet were designed to replicate the trigger on
top of an unimportant part of its output. When the compiled layers are
subsequently linked together, ImpNet would be chained between them, and still
effective on the overall model.

Further difficulty would be added for the attacker if different
compilers were used for each layer of the model, as each compiler must
be infected for the attack to succeed. However, we cannot recommend this
as a strategy for defending against ImpNet. Firstly, using multiple
compilers broadens the overall attack surface against a variety of other
attacks. Further, even if only the compiler for the first layer is
infected, this would still be sufficient for ImpNet to wreak havoc.

\vspace{4cm}
\balance
\hypertarget{conclusion}{%
\section{Conclusion}\label{conclusion}}

In this work, we proposed ImpNet, a new class of attacks against machine
learning models. ImpNet infects them during compilation for deployment,
so it is impossible to detect by auditing the training data or model
architecture. ImpNet does not touch the outputs when the input is clean,
and as its triggers are both \emph{imperceptible} and
\emph{high-entropy}, they are unlikely to be found by a defender.

We examined existing defences against ML backdoors, and found that
ImpNet cannot be reliably detected, although there are some defences
that might mitigate its effectiveness -- for a computational
price. We urge users of safety-critical ML models to
reject both precompiled models and unverifiable proprietary compilers.
We urge ML compiler teams to keep a tight watch on their
source code, even if this means it is no longer possible to support
every backend. Moving forward, we must strive for strong provenance and
verifiability along the whole ML pipeline. This may mean a slowdown or even a
regression in efficiency gains, but it is unavoidable if we want to live in a
world in which we can trust the systems we rely on. If not, we open the door to
powerful and covert attacks like ImpNet.

\FloatBarrier

\nocite{li2022backdoor}
\bibliographystyle{IEEEtranN}
\bibliography{bibliography}

\newpage

\FloatBarrier

\begin{table*}[p]
\hypertarget{app:detailed}{%
\appendix
\subsection{Detailed explanation of the elements of \rfig{pipeline}}\label{app:detailed}}
\vspace{1cm}
\centering
\caption{Detailed explanations of the inspection points in
\rfig{pipeline}.}
\vspace{0.2cm}
\begin{center}
\adjustbox{width=0.9\textwidth}{
\begin{tabular}{ll}
\toprule
Inspection point & Detailed explanation \\
\midrule

1 & The original data that is collected for use in training and
validation \\
2 & The original data, but with useless datapoints, outliers, poorly
labeled \\
& data, and so on removed. \\
3 & Data that is to be used for testing and validating the model. \\
4 & Data that is to be used for training the model. \\
5 & Data that is to be used for testing and validating the model,
after \\
& preprocessing. For example, after rotation and/or color jittering. \\
6 & Data that is to be used for training the model, after
preprocessing. \\
& For example, after rotation and/or color jittering. \\
7 & Data that is to be used for training the model, after sampling
e.g.~to \\
& separate it into batches for stochastic gradient descent. \\
8 & The hyperparameters of the model, for example the number and type
of \\
& layers. \\
9 & The actual architecture of the model, specified in a library such
as \\
& PyTorch or Tensorflow. \\
10 & The source code of the compiler which is used to compile the model
for \\
& deployment. \\
11 & The model represented in the compiler's Graph IR, for example
TVM's \\
& Relay. \\
12 & The model represented in the compiler's Operator IR, for example
TVM's \\
& TIR. \\
13 & The model represented in the IR of the backend the compiler is
using, \\
& for example LLVM or CUDA. \\
14 & The initial weights that are used at the start of training. \\
15 & The hyperparameters of training, for example learning rate,
dropout, \\
& rate, configuration and choice of optimizer, and so on. \\
16 & The weights after the model has been trained. \\
17 & The weights after optimization, usually for efficiency, for
example \\
& after quantization. \\
18 & The hardware which the model will run on. \\
19 & The runtime which interprets or JIT-compiles the Graph IR. \\
20 & The model represented as a graph which the runtime can interpret.
This \\
& might only be superficially different to (11) \\
21 & The machine code that is generated ahead of time by the
compiler. \\
22 & The operating system that is running the model. \\
23 & The inputs to the model. \\
24 & The model, viewed as a blackbox, i.e.~when only the inputs and
outputs \\
& can be observed. \\
\bottomrule

\end{tabular}
}
\end{center}
\end{table*}

% You do NOT want to edit this manually. make-md.sh has messed with it, yada yada
\begin{table*}[p]
\centering
\caption{Detailed explanations of the backdoor insertion points in
\rfig{pipeline}.}
\vspace{0.2cm}
\begin{center}
\adjustbox{width=0.9\textwidth}{
\begin{tabular}{ll}
\toprule
Insertion point & Detailed explanation \\
\midrule

A & The original data. \\
B & The process of removing useless datapoints, outliers, poorly
labeled \\
& data, and so on. \\
C & The process of splitting the entire dataset into training data
and \\
& test/validation data. \\
D & The preprocessing of the test/validation dataset, e.g.~random
rotation \\
& and color jittering. \\
E & The preprocessing of the training dataset, e.g.~random rotation
and \\
& color jittering. \\
F & The sampling of the training dataset, e.g.~to separate it into
batches \\
& for stochastic gradient descent. \\
G & The design of the model architecture, e.g.~deciding on \\
& hyperparameters, and implementing in a particular framework. \\
H & The translation of the model architecture from a framework's \\
& representation to a Graph IR. \\
I & The optimisation of the Graph IR, and the lowering to Operator
IR. \\
& These lines between these two processes are not always distinct. \\
J & The optimisation of the Operator IR, and the lowering to Backend
IR. \\
& These lines between these two processes are not always distinct. \\
K & The compilation of the Backend IR to machine code, e.g.~by LLVM. \\
L & The translation of the model from Graph IR to the Runtime Graph.
This \\
& may only be superficial. \\
M & The initial weights that are used at the start of training. \\
N & The hyperparameters of training, for example learning rate,
dropout, \\
& rate, configuration and choice of optimizer, and so on. \\
O & The training itself. \\
P & The weights after the model has been trained. \\
Q & The optimization of the weights, usually for efficiency, for
example \\
& quantization. \\
R & The weights after optimization, usually for efficiency, for
example \\
& after quantization. \\
S & The hardware which the model will run on. \\
T & The runtime which interprets or JIT-compiles the Graph IR. \\
U & The model represented as a graph which the runtime can interpret.
This \\
& might only be superficially different from (11) \\
V & The machine code that is generated ahead of time by the compiler. \\
W & The operating system that is running the model. \\
X & The inputs to the model. \\
\bottomrule

\end{tabular}
}
\end{center}
\end{table*}

\end{document}